\documentclass[11pt,a4paper]{article}
\usepackage[hyperref]{acl2018}
\usepackage{times}
\usepackage{latexsym}
\usepackage{url}

\usepackage{amsmath}
\usepackage{booktabs}
\usepackage{comment}
\usepackage{graphicx}
\usepackage[toc,page]{appendix}

\aclfinalcopy 

\newcommand{\appref}[1]{Appendix~\ref{sec:#1}}

\newcommand{\isection}[2]{\section{#1}\label{sec:#2}}

\title{LSTMs Exploit Linguistic Attributes of Data}

\author{ 
    Nelson F. Liu$^{\spadesuit\diamondsuit}$ \quad
	\bf Omer Levy$^\spadesuit$ \quad
	Roy Schwartz$^{\spadesuit\clubsuit}$ \quad 
	{\bf Chenhao Tan}$^\heartsuit$ \quad
	{\bf Noah A.~Smith}$^\spadesuit$ \\
    $^\spadesuit$Paul G. Allen School of Computer Science \& Engineering, University of Washington, Seattle, WA, USA \\
	$^\diamondsuit$Department of Linguistics, University of Washington, Seattle, WA, USA \\
	$^\clubsuit$Allen Institute for Artificial Intelligence, Seattle, WA, USA \\
    $^\heartsuit$Department of Computer Science, University of Colorado, Boulder, CO, USA \\
	{\tt \{nfliu,omerlevy,roysch,nasmith\}@cs.washington.edu,} \\  
	{\tt chenhao.tan@colorado.edu}
}

\date{}

\begin{document}
\maketitle
\begin{abstract}
  While recurrent neural networks have found success in a variety of natural language processing applications, they are general models of sequential data. We investigate how the properties of natural language data affect an LSTM's ability to learn a nonlinguistic task: recalling elements from its input. We find that models trained on natural language data are able to recall tokens from much longer sequences than models trained  on non-language sequential data. Furthermore, we show that the LSTM learns to solve the memorization task by explicitly using a subset of its neurons to count timesteps in the input. We hypothesize that the patterns and structure in natural language data enable LSTMs to learn by providing approximate ways of reducing loss, but understanding the effect of different training data on the learnability of LSTMs remains an open question.
\end{abstract}

\section{Introduction}

Recurrent neural networks (RNNs;  \citealp{Elman1990FindingSI}), especially variants with gating mechanisms such as long short-term memory units (LSTM; \citealp{Hochreiter1997LongSM}) and gated recurrent units (GRU;  \citealp{Cho2014LearningPR}), have significantly advanced the state of the art in many NLP tasks \citep[among others]{Mikolov2010RecurrentNN,Vinyals2015GrammarAA,Bahdanau2014NeuralMT}. However, RNNs are general models of sequential data; they are not explicitly designed to capture the unique properties of language that distinguish it from generic time series data.

In this work, we probe how linguistic properties such as the hierarchical structure of language \citep{everaert2015structures}, the dependencies between tokens, and the Zipfian distribution of token frequencies \citep{zipf1935psycho} affect the ability of LSTMs to learn. To do this, we define a simple memorization task where the objective is to recall the identity of the token that occurred a fixed number of timesteps in the past, within a fixed-length input. Although the task itself is not linguistic, we use it because (1) it is a generic operation that might form part of a more complex function on arbitrary sequential data, and (2) its simplicity allows us to unfold the mechanism in the trained RNNs. 
 
To study how linguistic properties of the training data affect an LSTM's ability to solve the memorization task, we consider several training regimens. 
In the first, we train on data sampled from a uniform distribution over a fixed vocabulary. 
In the second, the token frequencies have a Zipfian distribution, but are otherwise independent of each other. 
In another, the token frequencies have a Zipfian distribution but we add Markovian dependencies to the data.
Lastly, we train the model on natural language sequences.
To ensure that the models truly memorize, we evaluate on uniform samples containing {\em only rare words}.\footnote{This distribution is adversarial with respect to the Zipfian and natural language training sets.}

We observe that LSTMs trained to perform the memorization task on natural language data or data with a Zipfian distribution are able to memorize from sequences of greater length than LSTMs trained on uniformly-sampled data. 
Interestingly, increasing the length of Markovian dependencies in the data does not significantly help LSTMs to learn the task.
We conclude that linguistic properties can help or even enable LSTMs to learn the memorization task. 
Why this is the case remains an open question, but we propose that the additional structure and patterns within natural language data provide additional noisy, approximate paths for the model to minimize its loss, thus offering more training signal than the uniform case, in which the only way to reduce the loss is to learn the memorization function.

We further inspect \emph{how} the LSTM solves the memorization task, and find that some hidden units count the number of inputs. \citet{shi2016neural} analyzed LSTM encoder-decoder translation models and found that similar counting neurons regulate the length of generated translations. Since LSTMs better memorize (and thus better count) on language data than on non-language data, and counting plays a role in encoder-decoder models, our work could also lead to improved training for sequence-to-sequence models in non-language applications (e.g., \citealp{schwaller2017found}).

\section{The Memorization Task}

To assess the ability of LSTMs to retain and use information, we propose a simple memorization task. The model is presented with a sequence of tokens and is trained to recall the identity of the middle token.\footnote{Or the $(\frac{n}{2} + 1)$th token if the sequence length $n$ is even.}
We predict the middle token since predicting items near the beginning or the end might enable the model to avoid processing long sequences (e.g., to perfectly memorize the last token, simply set the forget gate to 0 and the input gate to 1).\footnote{We experimented with predicting tokens at a range of positions, and our results are not sensitive to the choice of predicting {\em exactly} the middle token.} All input sequences at train and test time are of equal length. To explore the effect of sequence length on LSTM task performance, we experiment with different input sequence lengths (10, 20, 40, 60, \ldots, 300).

\section{Experimental Setup}

We modify the linguistic properties of the training data and observe the effects on model performance.
Further details are found in \appref{Experiments}, and we release code for reproducing our results.\footnote{\url{http://nelsonliu.me/papers/lstms-exploit-linguistic-attributes/}}

\paragraph{Model.} 

We train an LSTM-based sequence prediction model to perform the memorization task. The model embeds input tokens with a randomly initialized embedding matrix. The embedded inputs are encoded by a single-layer LSTM and the final hidden state is passed through a linear projection to produce a probability distribution over the vocabulary. 

Our goal is to evaluate the memorization ability of the LSTM, so we freeze the weights of the embedding matrix and the linear output projection during training. This forces the model to rely on the LSTM parameters (the only trainable weights), since it cannot gain an advantage in the task by shifting words favorably in either the (random) input or output embedding vector spaces.
We also tie the weights of the embeddings and output projection so the LSTM can focus on memorizing the timestep of interest rather than also transforming input vectors to the output embedding space.\footnote{Tying these weights constrains the embedding size to always equal the LSTM hidden state size.}
Finally, to examine the effect of model capacity on memorization ability, we experiment with different hidden state size values. 

\paragraph{Datasets.} 

We experiment with several distributions of training data for the memorization task. In all cases, a 10K vocabulary is used.

\begin{itemize}
  \item In the \textit{uniform} setup, each token in the training dataset is randomly sampled from a uniform distribution over the vocabulary.
  
  \item In the \textit{unigram} setup, we modify the \textit{uniform} data by integrating the Zipfian token frequencies found in natural language data. The input sequences are taken from a modified version of the Penn Treebank \citep{marcus1993building} with randomly permuted tokens.
  
  \item In the \textit{5gram}, \textit{10gram}, and \textit{50gram} settings, we seek to augment the \textit{unigram} setting with Markovian dependencies. We generate the dataset by grouping the tokens of the Penn Treebank into 5, 10, or 50-length chunks and randomly permuting these chunks.
  
  \item In the \textit{language} setup, we assess the effect of using real language. The input sequences here are taken from the Penn Treebank, and thus this setup further extends the \textit{5gram}, \textit{10gram}, and \textit{50gram} datasets by adding the remaining structural properties of natural language.
\end{itemize}

We evaluate each model on a test set of uniformly sampled tokens from the 100 rarest words in the vocabulary. 
This evaluation setup ensures that, regardless of the data distribution the models were trained on, they are forced to generalize in order to perform well on the test set.
For instance, in a test on data with a Zipfian token distribution, the model may do well by simply exploiting the training distribution (e.g., by ignoring the long tail of rare words).

\section{Results}

We first observe that, in every case, the LSTM is able to perform the task perfectly (or nearly so), up to some input sequence length threshold. 
Once the input sequence length exceeds this threshold, performance drops rapidly.

\begin{figure}[t]
  \centering
  \includegraphics[width=\linewidth]{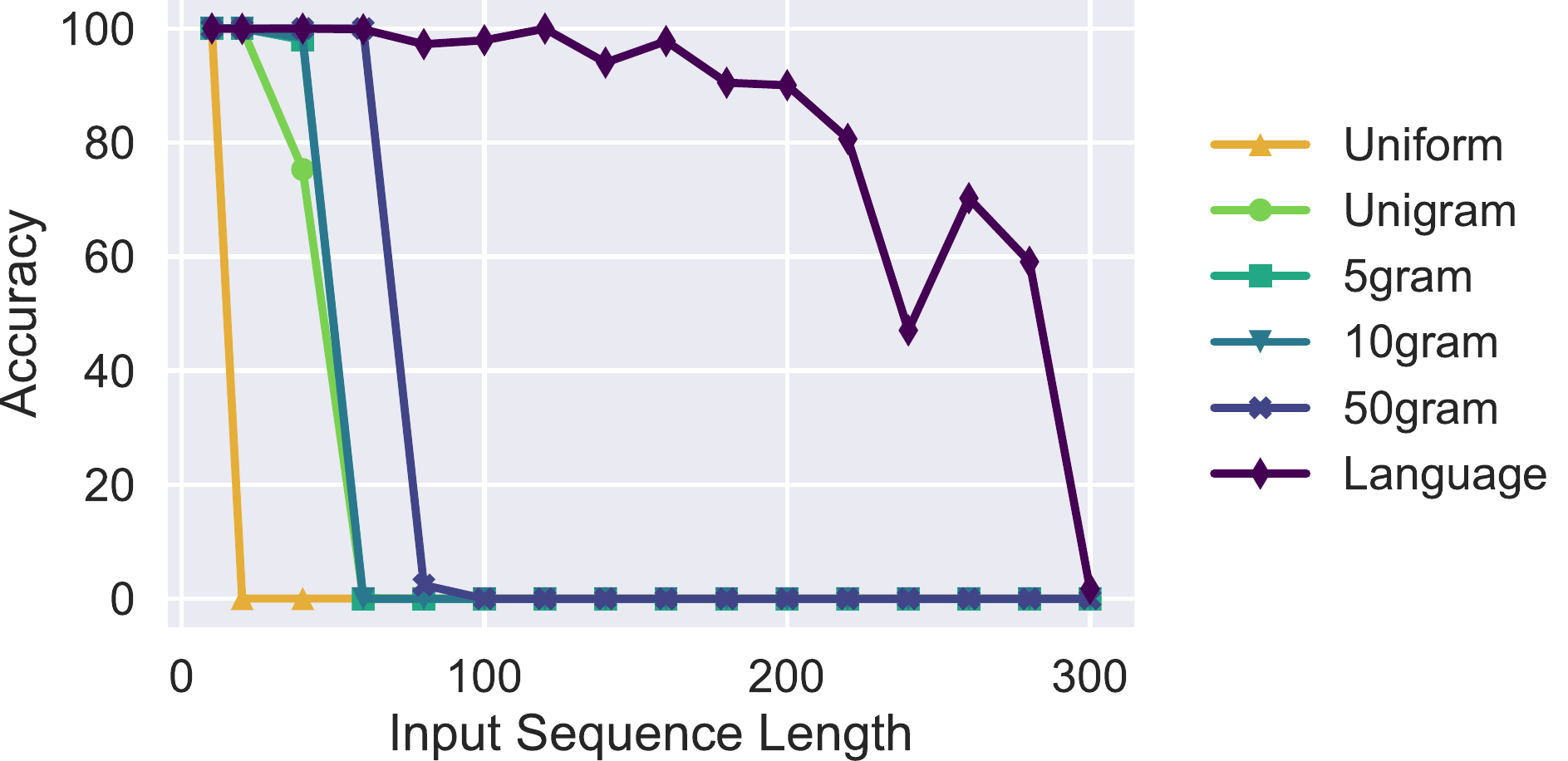}
  \caption{
   Test set accuracy of LSTMs with 50 hidden units trained on the \textit{uniform}, \textit{$\ast$gram}, and \textit{language} datasets with various input sequence lengths. \textit{5gram} and \textit{10gram} perform nearly identically, so the differences may not be apparent in the figure. \textit{unigram} accuracy plateaus to 0, and \textit{uniform} accuracy plateaus to $\approx$0.01\% (random baseline). Best viewed in color.}
  \label{fig:datasets_comparison}
\end{figure}

\paragraph{How does the training data distribution affect performance on the memorization task?} Figure~\ref{fig:datasets_comparison} compares memorization performance of an LSTM with 50 hidden units on various input sequence lengths when training on each of the datasets. Recall that the test set of only rare words is fixed for each length, regardless of the training data. When trained on the \textit{uniform} dataset, the model is perfect up to length 10, but does no better than the random baseline with lengths above 10.  Training on the \textit{unigram} setting enables the model to memorize from longer sequences (up to 20), but it begins to fail with input sequences of length 40; evaluation accuracy quickly falls to 0.\footnote{Manual inspection of the trained models reveals that they predict the most frequent words in the corpus. Since the evaluation set has only the 100 rarest types, performance (0\% accuracy) is actually worse than in the \textit{uniform} setting.} 
Adding Markovian dependencies to the \textit{unigram} dataset leads to small improvements, enabling the LSTM to successfully learn on inputs of up to length 40 (in the case of \textit{5gram} and \textit{10gram}) and inputs of up to length 60 (in the case of \textit{50gram}).
Lastly, training on \textit{language} significantly improves model performance, and it is able to perfectly memorize with input sequences of up to 160 tokens before any significant degradation. These results clearly indicate that training on data with linguistic properties helps the LSTM learn the non-linguistic task of memorization, even though the test set has an adversarial non-linguistic distribution.

\begin{figure}[t]
  \centering
  \includegraphics[width=\linewidth]{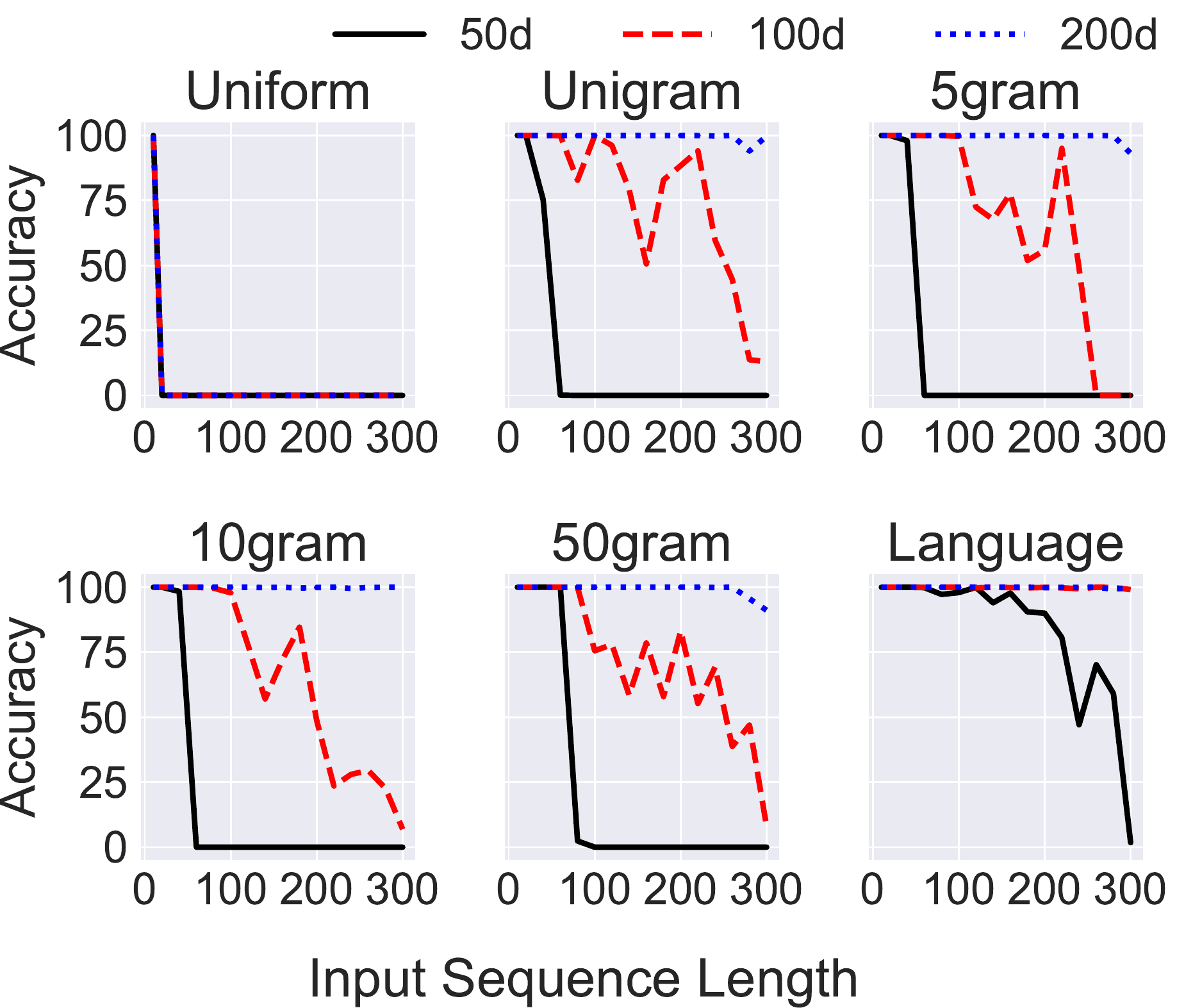}
  \caption{
  Test set accuracy of LSTMs with 50, 100 or 200 hidden units trained on each dataset with various input sequence lengths.}
  \label{fig:hidden_units_comparison}
\end{figure}

\paragraph{How does adding hidden units affect memorization performance?} Figure~\ref{fig:hidden_units_comparison} compares memorization performance on each dataset for LSTMs with 50, 100, and 200 hidden units. 
When training on the \textit{uniform} dataset, increasing the number of LSTM hidden units (and thus also the embedding size) to 100 or 200 does not help it memorize longer sequences. Indeed, even at 400 and 800 we saw no improvement (not shown in Figure~\ref{fig:hidden_units_comparison}).
When training on any of the other datasets, adding more hidden units eventually leads to perfect memorization for all tested input sequence lengths.
We take these results as a suggestion that successful learning for this task requires sufficiently high capacity (dimensionality in the LSTM). The capacity need is diminished  when the training data is linguistic, but LSTMs trained on the \textit{uniform} set cannot learn the memorization task even given high capacity.

\section{Analysis}

Throughout this section, we analyze an LSTM with 100 hidden units trained with the \textit{language} setting with an input sequence length of 300. This setting is a somewhat closer simulation of current NLP models, since it is trained on real language and recalls perfectly with input sequence lengths of 300 (the most difficult setting tested).

\paragraph{How do LSTMs solve the memorization task?}

A simple way to solve the memorization task is by counting. 
Since all of the input sequences are of equal length and the timestep to predict is constant throughout training and testing, a successful learner could maintain a counter from the start of the input to the position of the token to be predicted (the middle item).
Then, it discards its previous cell state, consumes the label's vector, and maintains this new cell state until the end of the sequence (i.e., by setting its forget gate near 1 and its input gate near 0). 

While LSTMs clearly have the expressive power needed to count and memorize, whether they can learn to do so from data is another matter.
Past work has demonstrated that the LSTMs in an encoder-decoder machine translation model learn to increment and decrement a counter to generate translations of proper length \citep{shi2016neural} and that representations produced by auto-encoding LSTMs contain information about the input sequence length \citep{adi2016fine}.
Our experiments isolate the counting aspect from other linguistic properties of translation and autoencoding (which may indeed be correlated with counting), and also test this ability with an adversarial test distribution and much longer input sequences. 

We adopt the method of \citet{shi2016neural} to investigate whether LSTMs solve the memorization task by learning to count. We identify the neurons that best predict timestep information by fitting a linear regression model to predict the number of inputs seen from the hidden unit activation. When evaluating on the test set, we observe that the LSTM cell state as a whole is very predictive of the timestep, with $\text{R}^2 =$ 0.998.

\begin{figure}[!t]
  \centering
  \includegraphics[width=\linewidth]{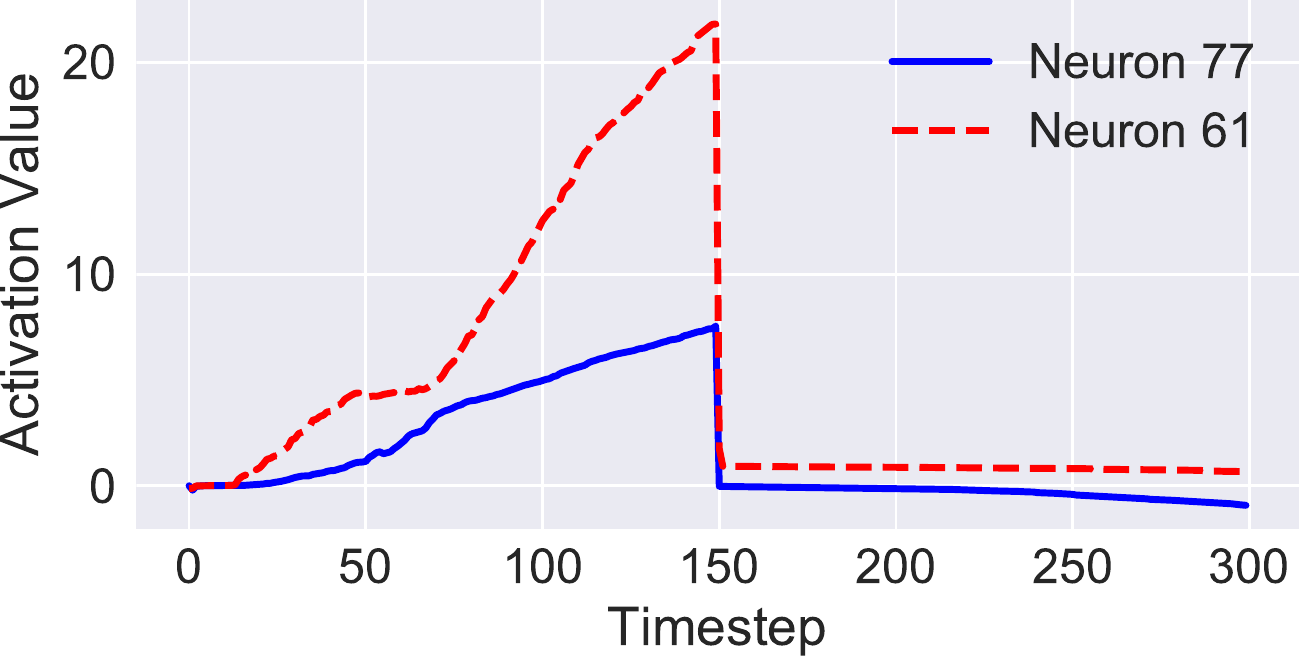}
  \caption{
  Activations of the neurons at indices 77 and 61 over time, showing counter-like behavior up to the target timestep to be remembered.
  }
  \label{fig:counting_neurons_activations}
\end{figure}

While no single neuron perfectly records the timestep, several of them are strongly correlated. In our model instance, neuron 77 has the highest correlation ($\text{R}^2 =$ 0.919), and neuron 61 is next ($\text{R}^2 =$ 0.901). The activations of these neurons over time for a random correctly classified test input linearly increase up to the target token, after which the activations falls to nearly 0 (Figure~\ref{fig:counting_neurons_activations}).

\paragraph{One hypothesis for why linguistic data helps.}

During training, the LSTM must: (1) determine what the objective is (here, ``remember the middle token'') and (2) adjust its weights to minimize loss. 
We observed that adding hidden units to LSTMs trained on \textit{unigram} or \textit{language} sets improves their ability to learn from long input sequences, but does not affect LSTMs trained on the \textit{uniform} dataset.
One explanation for this disparity is that LSTMs trained on \textit{uniform} data are simply not learning what the task is---they do not realize that the label always matches the  token in the middle of the input sequence, and thus they cannot properly optimize for the task, even with more hidden units. 
On the other hand, models trained on \textit{unigram} or \textit{language} can determine that the label is always the middle token, and can thus learn the task.
Minimizing training loss ought to be easier with more parameters, so adding hidden units to LSTMs trained on data with linguistic attributes increases the length of input sequences that they can learn from. 

But why might LSTMs trained on data with linguistic attributes be able to effectively learn the task for long input sequences, whereas LSTMs trained on the \textit{uniform} dataset cannot?  We conjecture that linguistic data offers more reasonable, if approximate, pathways to loss minimization, such as counting frequent words or phrases.
In the \textit{uniform} setting, the model has only one path to success: true memorization, and it cannot find an effective way to reduce the loss.
In other words, linguistic structure and the patterns of language may provide additional signals that correlate with the label and facilitate learning the memorization task.

\begin{figure}[!t]
  \centering
  \includegraphics[width=\linewidth]{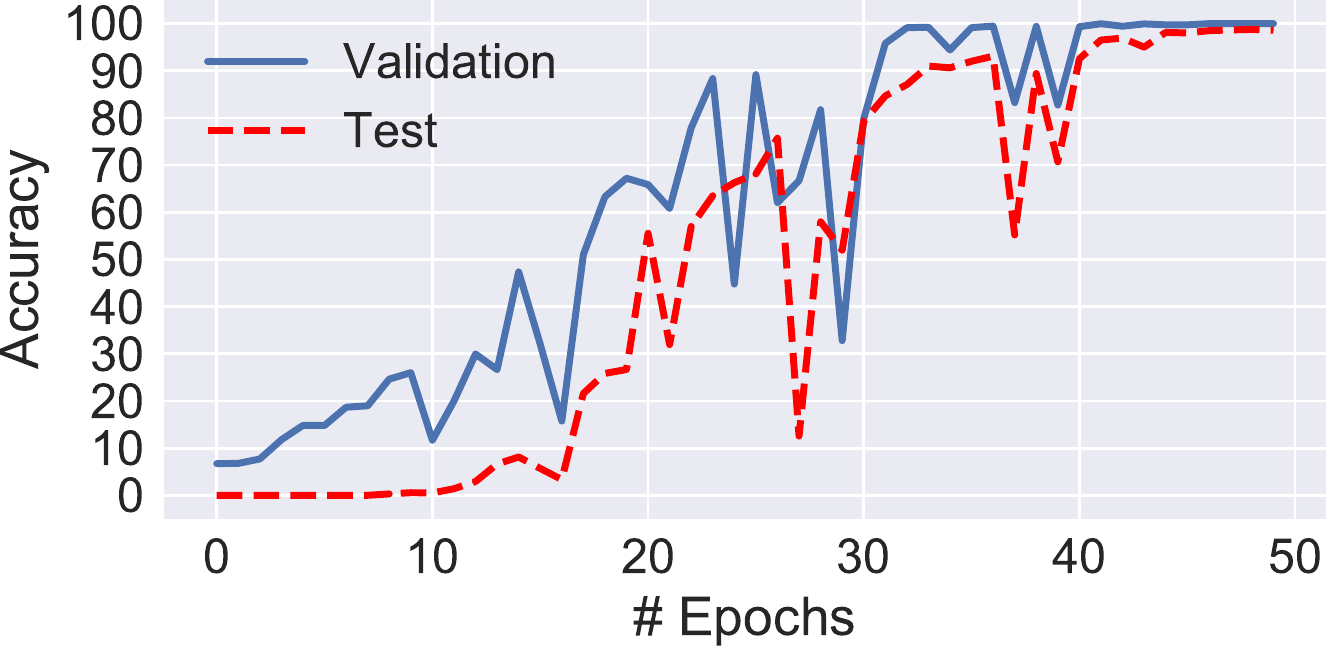}
  \caption{
  Model validation and test accuracy over time during training. Validation improves faster than test, indicating that the model exploits linguistic properties of the data during training.}
  \label{fig:validation_vs_test_convergence}
\end{figure}

Figure~\ref{fig:validation_vs_test_convergence} shows that models trained on the \textit{unigram} and \textit{language} datasets converge to high validation accuracy faster than high test accuracy. This suggests that models trained on data with linguistic attributes first learn to do well on the training data by exploiting the properties of language and not truly memorizing. Perhaps the model generalizes to actually recalling the target token later, as it refines itself with examples from the long tail of infrequent tokens. 

Figure~\ref{fig:validation_vs_test_convergence} may show this shift from exploiting linguistic properties to true memorization. The validation and test accuracy curves are quite synchronized from epoch 37 onward, indicating that the model's updates affect both sets identically. The model clearly learns a strategy that works well on both datasets, which strongly suggests that it has learned to memorize. In addition, when the model begins to move toward true memorization, we'd expect validation accuracy to momentarily falter as it moves away from the crutches of linguistic features---this may be the dip at around epoch 35 from perfect validation accuracy to around 95\% accuracy. 

\section{Related Work}

To our knowledge, this work is the first to study how linguistic properties in training data affect the ability of LSTMs to learn a general, non-linguistic, sequence processing task.

Previous studies have sought to better understand the empirical capabilities of LSTMs trained on natural language data. \citet{linzen2016assessing} measured the ability of LSTMs to learn syntactic long range dependencies commonly found in language, and \citet{gulordava2018colorless} provide evidence that LSTMs can learn the hierarchical structure of language. \citet{blevins2018deep} show that the internal representations of LSTMs encode syntactic information, even when trained without explicit syntactic supervision.

Also related is the work of \citet{weiss2018practical}, who demonstrate that LSTMs are able to count infinitely, since their cell states are unbounded, while GRUs cannot count infinitely since the activations are constrained to a finite range. One avenue of future work could compare the performance of LSTMs and GRUs on the memorization task.

Past studies have also investigated what information RNNs encode by directly examining hidden unit activations \citep[among others]{karpathy2015visualizing, li2016visualizing, shi2016neural} and by training an auxiliary classifier to predict various properties of interest from hidden state vectors \citep[among others]{shi2016does,adi2016fine,belinkov2017neural}.

\section{Conclusion}

In this work, we examine how linguistic attributes in training data can affect an LSTM's ability to learn a simple memorization task. 
We find that LSTMs trained on uniformly sampled data are only able to learn the task with the sequence length of 10, whereas LSTMs trained with language data are able to learn on sequences of up to 300 tokens. 

We further investigate how the LSTM learns to solve the task, and find that it uses a subset of its hidden units to track timestep information. It is still an open question why LSTMs trained on linguistic data are able to learn the task whereas LSTMs trained on uniformly sampled data cannot; based on our observations, we hypothesize that the additional patterns and structure in language-based data may provide the model with 
approximate paths of loss minimization, and improve LSTM trainability as a result.

\section*{Acknowledgments}

We thank the ARK as well as the anonymous reviewers for their valuable feedback. NL is supported by a Washington Research Foundation Fellowship and a Barry M. Goldwater Scholarship. This work was supported in part by a hardware gift from NVIDIA Corporation and a UW High Performance Computing Club Cloud Credit Award.
\bibliography{lstm_memorization}
\bibliographystyle{acl_natbib}

\clearpage

\begin{appendices}

\isection{Experimental Setup Details}{Experiments}

\paragraph{Penn Treebank Processing} Our experiments use a preprocessed version of the Penn Treebank commonly used in the language modeling community and first introduced by \citet{mikolov2011extensions}. This dataset has 10K types, hence why we use this vocabulary size for all experiments. We generate examples by concatenating the sentences together and taking subsequences of the desired input sequence length.

\paragraph{Training} The model is trained end-to-end to directly predict the tokens at a particular timestep in the past; it is optimized with Adam \citep{Kingma2015Adam} with an initial learning rate of 0.001, which is halved whenever the validation dataset (a held-out portion of the training dataset) loss fails to improve for three consecutive epochs. The model is trained for a maximum of 240 epochs or until it converges to perfect validation performance. We do not use dropout; we included it in initial experiments, but it severely hampered model performance and does not make much sense for a task where the goal is to explicitly memorize. We ran each experiment three times with different random seeds and evaluate the model with the highest validation accuracy on the test set. We take the best since we are interested in whether the LSTMs \textit{can} be trained for the task.

\end{appendices}

\end{document}